\documentclass[journal]{IEEEtran}
\usepackage{times}
\usepackage{amssymb}
\usepackage{helvet}
\usepackage{mathrsfs}
\usepackage{url}
\usepackage{courier}
\usepackage{booktabs}
\usepackage{threeparttable}
\usepackage{amsmath}
\usepackage{graphicx}
\usepackage{url}	
\usepackage{subfig}
\usepackage{overpic}
\usepackage{color}
\usepackage[english]{babel}
\usepackage[T1]{fontenc}
\usepackage[utf8]{inputenc}
\usepackage{multirow}
\usepackage{diagbox}
\usepackage{algorithm}  
\usepackage{algpseudocode} 
\usepackage[square,sort,comma,numbers,compress]{natbib}
\usepackage{varwidth}
\usepackage{siunitx} 
\usepackage{xpatch}
\xpatchcmd{\citet}{\begingroup}{\begingroup\em}{}{}
\xpatchcmd{\citeauthor}{\begingroup}{\begingroup\em}{}{}

\begin{document}

\title{Transformation-consistent Self-ensembling Model for Semi-supervised Medical Image Segmentation}

\author{Xiaomeng Li,
	Lequan Yu,~\IEEEmembership{Member,~IEEE,}
	Hao Chen,~\IEEEmembership{Member,~IEEE,}\\
	Chi-Wing Fu,~\IEEEmembership{Member,~IEEE,}
	Lei Xing,
	and Pheng-Ann Heng,~\IEEEmembership{Senior~Member,~IEEE}
	
	\thanks{X. Li, L. Yu, H. Chen, C.-W. Fu and P.-A. Heng are with the Department of Computer Science and Engineering, The Chinese University of Hong Kong, Hong Kong (e-mail: xmli@cse.cuhk.edu.hk; lqyu@cse.cuhk.edu.hk; hchen@cse.cuhk.edu.hk; cwfu@cse.cuhk.edu.hk; pheng@cse.cuhk.edu.hk).} 
	\thanks{X. Li, L. Yu and L. Xing are also with the Department of Radiation Oncology, Stanford University, Stanford, CA 94305 USA (lei@stanford.edu).}
}

%

\markboth{}%
{Shell \MakeLowercase{\textit{et al.}}: Bare Demo of IEEEtran.cls for IEEE Journals}
%



\def\ie{\emph{i.e.}}
\def\eg{\emph{e.g.}}
\def\etal{{\em et al.}}
\def\etc{{\em etc.}}
\renewcommand{\algorithmicrequire}{\textbf{Input:}}
\newcommand{\para}[1]{\vspace{.05in}\noindent\textbf{#1}}

\newcommand{\TODO}[1]{{\color{red}{[TODO: #1]}}}
\newcommand{\xmli}[1]{{\color{magenta}{[XM:#1]}}}
\newcommand{\ylq}[1]{{\color{green}{[LQ:#1]}}}
\newcommand{\revise}[1]{{\color{black}{#1}}}
\newcommand{\reviselast}[1]{{\color{black}{#1}}}
\newcommand{\reviseagain}[1]{{\color{black}{#1}}}

\maketitle



\IEEEpeerreviewmaketitle

\begin{abstract}
\reviseagain{A common shortfall of supervised deep learning for medical imaging is the lack of labeled data, which is often expensive and time-consuming to collect. 
This paper presents a new semi-supervised method for medical image segmentation, where the network is optimized by a weighted combination of a common supervised loss only for the labeled inputs} and a regularization loss for both the labeled and unlabeled data.
To utilize the unlabeled data, our method encourages consistent predictions of the network-in-training for the same input under different perturbations.
\reviseagain{With the semi-supervised segmentation tasks}, we introduce a transformation-consistent strategy in the self-ensembling model to enhance the regularization effect for pixel-level predictions. To further improve the regularization effects, \reviseagain{we extend the transformation in a more generalized form including scaling and optimize the consistency loss with a teacher model, which is an averaging of the student model weights.}
We extensively validated the proposed semi-supervised method on three typical yet challenging medical image segmentation tasks: (i) \reviseagain{skin lesion segmentation from dermoscopy images in the International Skin Imaging Collaboration (ISIC) 2017 dataset,} (ii) optic disc segmentation from fundus images in the Retinal Fundus Glaucoma Challenge (REFUGE) dataset, and (iii) liver segmentation from volumetric CT scans in the Liver Tumor Segmentation Challenge (LiTS) dataset.
\reviseagain{Compared to state-of-the-art, our method shows superior performance on the challenging 2D/3D medical images, demonstrating the effectiveness of our semi-supervised method for medical image segmentation.}
\end{abstract}
\begin{IEEEkeywords}
semi-supervised learning, self-ensembling, skin lesion segmentation, optic disc segmentation, liver segmentation.
\end{IEEEkeywords}
\section{Introduction}
\begin{figure}[t]
	\centering
	\includegraphics[width=0.85\linewidth]{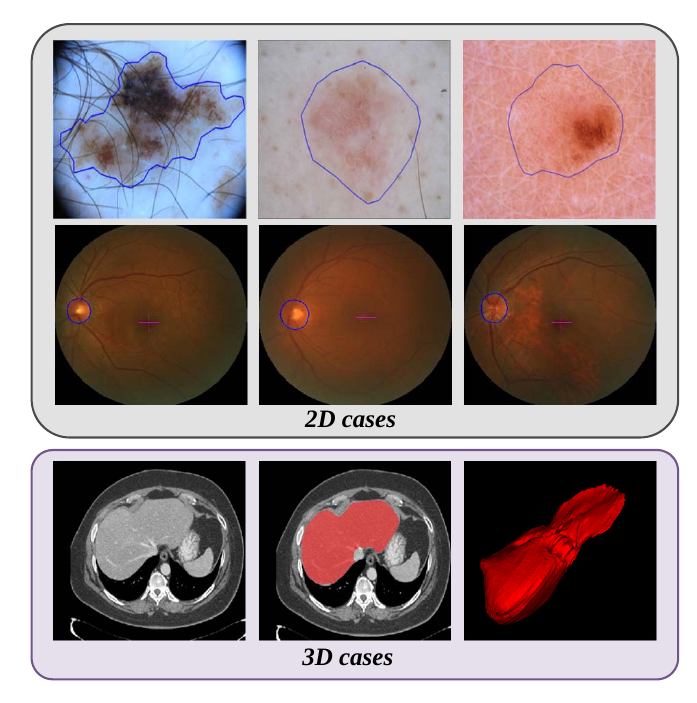}
	\vspace*{-2mm}
	\caption{
		\reviseagain{Three common medical image segmentation tasks. The first, second, and third rows show the skin lesion in the dermoscopy image, the optical disc in retinal fundus images, and liver segmentations from CT scans. Blue color denotes the structure boundary and red color represents the liver.}}
	\label{fig:intro}\centering
	\vspace*{-4.5mm}
\end{figure}

\IEEEPARstart{S}{egmenting} anatomical structural or abnormal regions from medical images, \reviselast{such as dermoscopy images, fundus images, and 3D computed tomography (CT) scans, is of great significance for clinical practice, especially for disease diagnosis and treatment planning}.
Recently, deep learning techniques have made impressive progress on semantic image segmentation tasks and become a popular choice in both computer vision and medical imaging community~\cite{nie2018strainet,mahmud2018applications}. 
The success of deep neural networks usually relies on the massive labeled dataset.
However, it is hard and expensive to obtain labeled data, \reviselast{notably in the medical imaging domain where only experts can provide reliable annotations~\cite{kohli2017medical}.}
For example, there are thousands of dermoscopy image records in the clinical center, but melanoma delineation by experienced dermatologists is very scarce; see Figure~\ref{fig:intro}. Such cases can also be observed in the optic disc segmentation from the retinal fundus images, and especially in liver segmentation from CT scans, where delineating organs from volumetric images in a slice-by-slice manner is very time-consuming and expensive. 
\reviseagain{The lack of the labeled data motivates the study of methods that can be trained with limited supervision, such as semi-supervised learning~\cite{zhou2018semi,sedai2017semi,cheplygina2018not}, weakly supervised learning~\cite{hu2018weakly,gondal2017weakly,feng2017discriminative}, and unsupervised domain adaptation~\cite{kamnitsas2017unsupervised,dong2018unsupervised,mahmood2018unsupervised}, etc. }
In this paper, we \reviseagain{focus on the semi-supervised} segmentation approaches, considering that it is relatively easy to acquire a large amount of unlabeled medical image data.

Semi-supervised learning aims to learn from a limited amount of labeled data and an arbitrary amount of unlabeled data, 
\reviselast{which is a fundamental, challenging problem and has a high impact on real-world clinical applications.}
The semi-supervised problem has been widely studied in medical image research community~\cite{you2011segmentation, portela2014semi,masood2015self,gu2017semi,jaisakthi2017automatic}.
Recent progress in semi-supervised learning for medical image segmentation has featured deep learning~\cite{bai2017semi,sedai2017semi,nie2018asdnet,perone2018deep,ganaye2018semi}.~\citeauthor{bai2017semi}~\cite{bai2017semi} present a semi-supervised deep learning model for cardiac MR image segmentation, where the segmented label maps from unlabeled data are incrementally added into the training set to refine the segmentation network.
Other semi-supervised learning methods are based on the recent techniques, such as variational autoencoder (VAE)~\cite{sedai2017semi} and generative adversarial network (GAN)~\cite{nie2018asdnet}.  
We tackle the semi-supervised segmentation problem from a different point of view. 
\reviselast{With the success of the self-ensembling model in the semi-supervised classification problem~\cite{laine2016temporal}}, we further advance the method to medical image segmentation tasks, including 2D cases and 3D cases.   

In this paper, we present \reviseagain{a new semi-supervised learning method based on the self-ensembling strategy for medical image segmentation.
The whole framework is trained with a weighted combination of supervised and unsupervised losses.
The supervised loss is designed to utilize the labeled data for accurate predictions.}
\revise{To leverage the unlabeled data, our self-ensembling method encourages a consistent prediction of the network for the same input under different regularizations, \eg, randomized Gaussian noise, network dropout, and randomized data transformation. 
\reviseagain{In particular, our method accounts for the challenging segmentation task, in which pixel-level classification is required to be predicted.} We observe that in the segmentation problem, if one transforms (\eg, rotates) the input image, the expected prediction should be transformed in the same manner. 
When the inputs of CNNs are rotated, the corresponding network predictions would not rotate in the same way~\cite{worrall2017harmonic}.~\reviseagain{In this regard, we take advantage of this property by introducing a transformation (\ie, rotation, flipping) consistent scheme at the input and output space of our network.}
Specifically, we design the unsupervised loss by minimizing the differences between the network predictions under different transformations of the same input.
\reviseagain{To further improve the regularization, we extend the transformation consistency regularization with the scaling operation and optimize the network under a scaling consistent scheme.}
\reviseagain{Also, we adopt a teacher model to evaluate images under perturbations to construct better targets.} 
We extensively evaluate our methods for semi-supervised medical image segmentation on three representative segmentation tasks, \ie, skin lesion segmentation from dermoscopy images, optic disc segmentation from retinal images, and liver segmentation from CT scans.
In summary, our semi-supervised method achieves significant improvements compared with the supervised baseline and also outperforms other semi-supervised segmentation methods.

The main contributions of this paper are:

\begin{itemize}
    
    \item We present a novel and effective semi-supervised method, namely TCSM\_v2, for medical image segmentation. Our method is flexible and can be easily applied to both 2D and 3D convolutional neural networks.  
    
    \reviseagain{
    \item We regularize unlabeled data with the transformation-consistent strategy and demonstrate effective semi-supervised medical image segmentation.}
    
    \item 
    Extensive experiments on three representative yet challenging medical image segmentation tasks, including 2D and 3D datasets, demonstrate the effectiveness of our semi-supervised method over other methods. 
   
    \item Our method excels other state-of-the-art methods and establishes a new record in the ISIC 2017 skin lesion segmentation dataset with the semi-supervised method.

\end{itemize}

This work extends our previous work TCSM~\cite{li2018semi} in three aspects. 
First, multi-scale inference is an effective technique utilized in many image recognition tasks~\cite{he2016deep,jegou2010product,he2015spatial}. \reviseagain{To enhance the regularization, we extend TCSM with more generalized transformation, such as random scaling.}~Through this, we utilize the unlabeled data to improve the regularization of the network. 
Second, our preliminary TCSM evaluates the inputs with perturbations on the same network. To avoid the misrecognition, \reviseagain{we incorporate a teacher model to construct better targets, where the teacher model is an exponential moving average of the student model.} 
Thirdly, we evaluate our method on three datasets, including the skin lesion dataset, retinal fundus dataset, and liver CT dataset. 
\reviseagain{Experiments on all three datasets show the effectiveness of our method over  existing methods for semi-supervised medical image segmentation. }
}

\begin{figure*}[!t]
	\centering
	\includegraphics[width=0.88\linewidth]{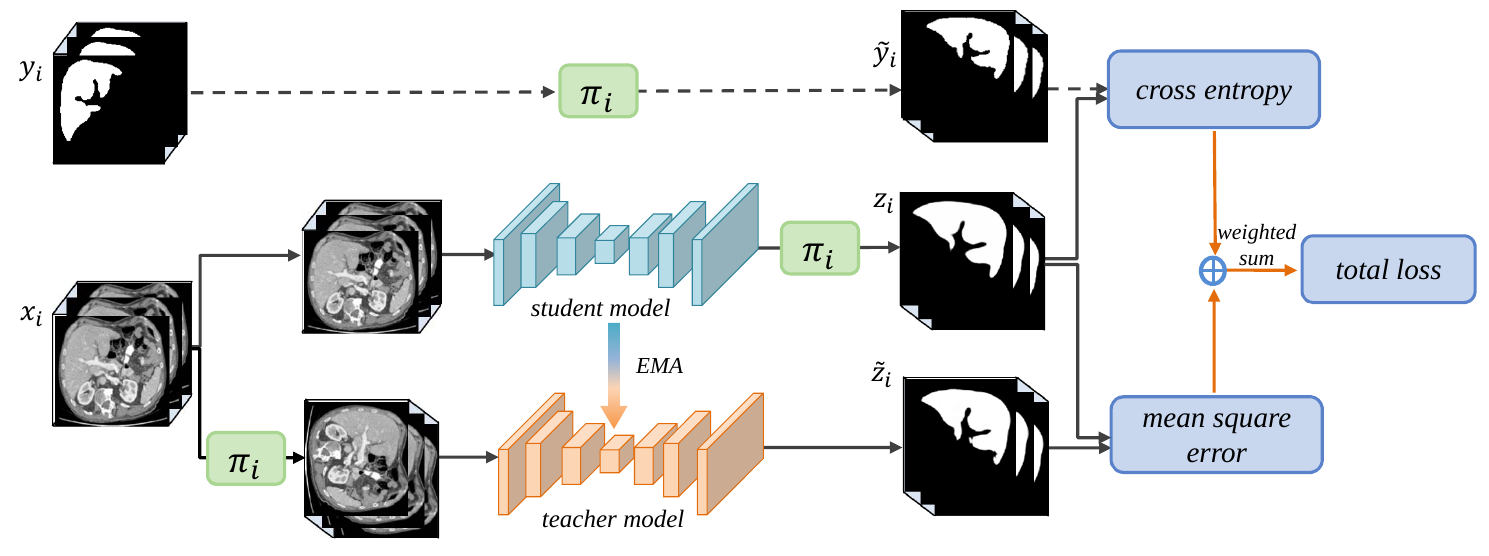}
	\vspace*{-1mm}
	\caption{\reviseagain{Our transformation-consistent self-ensembling model (TCSM\_v2) for semi-supervised medical image segmentation (we use liver CT scans as  example). }
	\reviseagain{The teacher and student models share the same architecture}, and the weight of the teacher model is the exponential moving average (EMA) of the student model.  
	The student model is trained by the total loss, which \reviseagain{is a weighted combination} of the cross-entropy loss on labeled data, and mean square error loss on both labeled and unlabeled data. 
	The model encourages the teacher and student models to be transformation-consistent by utilizing the unlabeled data. 
	$\pi_i$ refers to the transformation-consistent regularization, including rotation, flipping, and scaling operations. }
	\label{fig:pipeline}\centering
	\vspace*{-3.5mm}
\end{figure*}

\section{Related Work} 

\noindent
{\em Semi-supervised segmentation for medical images\/}.
Early semi-supervised works segment medical images mainly using hand-crafted features~\cite{you2011segmentation,portela2014semi,masood2015self,gu2017semi,jaisakthi2017automatic}.
\reviseagain{\citeauthor{you2011segmentation}~\cite{you2011segmentation} combined radial projection and self-training learning to improve the segmentation of retinal vessel from fundus image.}
\citeauthor{portela2014semi}~\cite{portela2014semi}~\reviselast{presented a clustering-based Gaussian mixture model to automatically segment brain MR images.}
Later on,~\citeauthor{gu2017semi}~\cite{gu2017semi} \reviselast{constructed forest oriented superpixels for vessel segmentation}.
For skin lesion segmentation,~\citeauthor{jaisakthi2017automatic}~\cite{jaisakthi2017automatic} explored K-means clustering and flood fill algorithm. 
\reviseagain{These semi-supervised methods are, however, based on hand-crafted features, which suffer from limited representation capacity.}

\reviseagain{Recent works for semi-supervised segmentation are mainly based on} deep learning. 
\reviseagain{An iterative method is proposed by~\citeauthor{bai2017semi}~\cite{bai2017semi} for cardiac segmentation from MR images}, where network parameters and segmentation masks for unlabeled data are alternatively updated. 
\reviselast{Generative model based semi-supervised approaches are also popular in the medical image analysis community~\cite{sedai2017semi,nie2018asdnet,zhang2017deep,chartsias2018factorised}}.~\citeauthor{sedai2017semi}~\cite{sedai2017semi} introduced a variational autoencoder (VAE) for optic cup segmentation from retinal fundus images.
They learned the feature embedding from unlabeled images using VAE, and then combined the feature embedding with the segmentation autoencoder trained on the labeled images for pixel-wise segmentation of the cup region. 
\reviseagain{To involve unlabeled data in training,~\citeauthor{nie2018asdnet}~\cite{nie2018asdnet} presented an attention-based GAN approach to select trustworthy regions of the unlabeled data to train the segmentation network.}
\reviseagain{Another GAN-based work~\cite{chartsias2018factorised} employed the cycle-consistency principle and performed experiments on cardiac MR image segmentation.}
More recently,~\citeauthor{ganaye2018semi}~\cite{ganaye2018semi} proposed a semi-supervised method for brain structures segmentation by taking advantage of the invariant nature and semantic constraint of anatomical structures.
\reviseagain{Multi-view co-training based methods~\cite{zhou2018semi,xia20183d} have also been explored on 3D medical data.
Differently, our method takes the advantage of transformation consistency and self-ensembling model, which is simple yet effective for medical image segmentation tasks.}

\vspace*{1.5mm}
\reviselast{
\noindent
{\em Transformation equivariant representation\/}.
\reviseagain{Next, we review equivariance representations, to which the transformation equivariance is encoded in the network to explore the network equivariance property~\cite{cohen2016group, pmlr-v48-dieleman16, worrall2017harmonic,li2018deeply}.}~\reviseagain{\citeauthor{cohen2016group}~\cite{cohen2016group} proposed a group equivariant neural network to improve} the network generalization, where equivariance to \ang{90}-rotations and dihedral flips are encoded by copying the transformed filters at different rotation-flip combinations.~Concurrently,~\citeauthor{pmlr-v48-dieleman16}~\cite{pmlr-v48-dieleman16} designed four different equivariance to preserve feature map transformations by \reviseagain{rotating the feature maps instead of the filters}. Recently,~\citeauthor{worrall2017harmonic}~\cite{worrall2017harmonic} restricted the filters to circular harmonics to achieve continuous \ang{360}-rotations equivariance. However, these works aim to encode equivariance into the network to improve its generalization capability, while our method aims to better utilize the unlabeled data in semi-supervised learning.}

\vspace*{1mm}
\noindent
{\em Medical image segmentation\/}.
\reviseagain{Early methods for medical image segmentation mainly focused on using thresholding~\cite{emre2013lesion}, statistical shape models~\cite{heimann2009statistical} and machine learning~\cite{he2012automatic,sadri2013segmentation,cheng2013superpixel,abramoff2007automated,chlebus2018automatic}, while recent ones are mainly deep-learning-based~\cite{cciccek20163d,ronneberger2015u,milletari2016v}.}
\reviseagain{Deep learning methods showed promising results on skin lesion segmentation, optic disc segmentation, and liver segmentation~\cite{yu2017automated,fu2018disc,tan2017segmentation,lu2017automatic,yang2017automatic}. 
\citeauthor{yu2017automated}~\cite{yu2017automated} explored the network depth property and developed a deep residual network for automatic skin lesion segmentation by stacking residual blocks to increase the network's representative capability.}
\citeauthor{yuan2017automatic}~\cite{yuan2017automatic} trained a 19-layer deep convolutional neural network in an end-to-end manner for skin lesion segmentation. 
As for optical disc segmentation, \reviseagain{\citeauthor{fu2018disc}~\cite{fu2018disc} presented an M-Net for joint OC and OD segmentation. }
\reviseagain{Also, a disc-aware network~\cite{fu2018disc} was designed for glaucoma screening by an ensemble of different feature streams of the network.} For liver segmentation, \citeauthor{chlebus2018automatic}~\cite{chlebus2018automatic} presented a cascaded FCN combined with hand-crafted features. 
\citeauthor{li2018h}~\cite{li2018h} presented a 2D-3D hybrid architecture for liver and tumor segmentation from CT images.~\reviseagain{Although these methods achieved good results, they are based on fully supervised learning, requiring massive pixel-wise annotations from experienced dermatologists or radiologists.}
\section{Method}
\reviseagain{Figure~\ref{fig:pipeline} overviews our transformation-consistent self-ensembling model (TCSM\_v2) for semi-supervised medical image segmentation.
First, we randomly sample $x_i$ raw data, including both the labeled and unlabeled cases from the training dataset, followed by performing random transformations on these images.  
Teacher and student models are formulated in our framework, where the student model is trained by the loss function and the teacher model is an average of consecutive student models. 
To train the student model, the transformed inputs are fed into the student model and the softmax output is compared with a one-hot label using classification cost (cross-entropy in Figure~\ref{fig:pipeline}) and with the teacher output using consistency cost (mean square error in Figure~\ref{fig:pipeline}). After the weights of the student model have been updated with gradient descent, the teacher model weights are updated as an exponential moving average of the student weights.
Hence, the label information is passed to the unlabeled data by constraining the model outputs to be consistent with the unlabeled data.}

\subsection{\revise{Mean Teacher Based Semi-supervised Framework} }
\reviseagain{To ease the description of our method, we first formulate the semi-supervised segmentation task. In the semi-supervised segmentation problem, the training set consists of $N$ inputs in total, including $M$ labeled inputs and $N-M$ unlabeled inputs. 
We denote the labeled set as $\mathcal{L} = \left \{ (x_i, y_i) \right \}_{i=1}^{M}$ and the unlabeled set as~$\mathcal{U} = \left \{x_i \right \}_{i=M+1}^{N}$. 
For the 2D images, $x_i \in \mathbb{R}^{H \times W \times 3}$ denotes the input image and $y_i \in \{0,1\}^{H \times W}$ is the ground-truth segmentation mask. 
For the 3D volumes, $x_i \in \mathbb{R}^{H \times W \times D}$ denotes the input volume and $y_i \in \{0,1\}^{H \times W \times D}$ is the ground-truth segmentation volume.}
The general semi-supervised segmentation learning tasks can be formulated to learn the network parameters $\theta$ by optimizing:
\begin{equation}
\min_{\theta} \sum_{i=1}^{M} l (f(x_i; \theta), y_i) + \lambda R(\theta, \mathcal{L}, \mathcal{U}),
\label{eq:semi}
\end{equation}
where $l$ is the supervised loss function, $R$ is the regularization (unsupervised) loss, and $f(\cdot)$ is the segmentation neural network \reviseagain{and $\theta$ denotes the model weights. $\lambda$ is a weighting factor that controls how strong the regularization is.
The first term in the loss function is trained by the cross-entropy loss, aiming at evaluating the correctness of network output on labeled inputs only.
The second term is optimized with the regularization loss, which utilizes both the labeled and unlabeled inputs. 

The key point of this semi-supervised learning is based on the \emph{smoothness} assumption,~\ie, data points close to each other in the image space are likely to be close in the label space~\cite{sajjadi2016regularization,laine2016temporal}.}
Specifically, these methods focus on improving the target quality using self-ensembling and exploring different perturbations, which include the input noise and the network dropout.
The network with the regularization loss encourages the predictions to be consistent and is expected to give better predictions. 
The regularization loss $R$ can be described as: 
\begin{equation}
R(\theta, \mathcal{L}, \mathcal{U}) =\sum_{i=1}^{N} \mathbb{E}_{\xi^{'}, \xi} \left \| f(x_i; \theta, \xi^{'}) - f(x_i;\theta, \xi) \right \| ^2,
\end{equation}
where $\xi$ and $\xi'$ denote to different regularization and perturbations of input data, respectively.
In our work, we share the same spirit as these methods by designing different perturbations for the input data. 
Specifically, we design the regularization term as a consistency loss to encourage smooth predictions for the same data under different regularization and perturbations (\eg, Gaussian noise, network dropout, and randomized data transformation).

\begin{figure}[!t]
	\centering
	\includegraphics[width=0.99\linewidth]{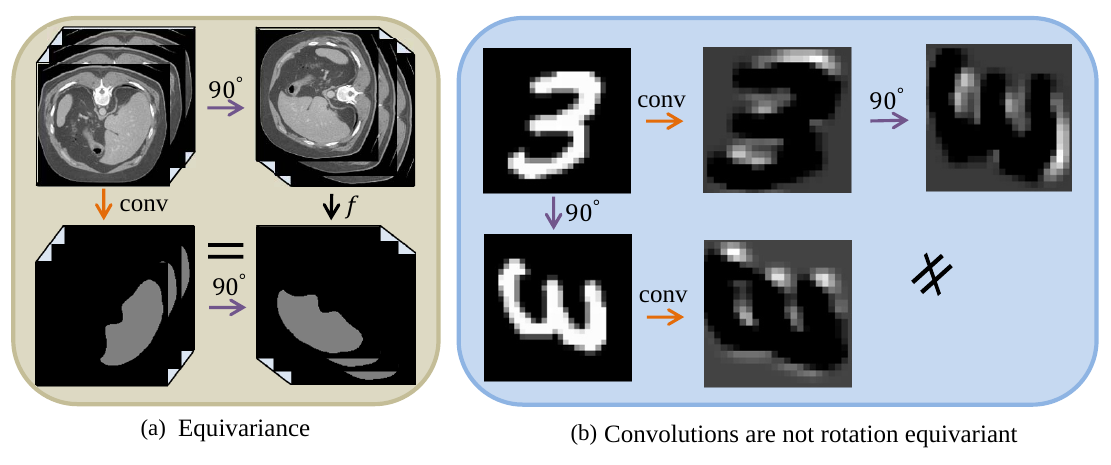}
	\vspace*{-5mm}
	\caption{(a) Segmentation is desired to be rotation equivariant. If the input image is rotated, the ground truth mask should be rotated in the same manner.
	(b) Convolutions are not rotation equivariant in general. If the input image is rotated, the generated output \reviseagain{is not the same with} the original output that rotated in the same manner.}
	\label{fig:method2}\centering
	\vspace*{-2mm}
\end{figure}

In the above, we evaluate the model twice to get two predictions under different perturbations.
In this case, the model assumes a dual role as a teacher and  as a student. 
As a student, it learns as before, while as a teacher, it generates targets, which are then used by itself as a student for learning. 
The model generates the targets by itself, thus it may be incorrect, especially when excessive weight is given to the generated targets. 
To construct better targets, we employ the mean teacher based framework~\cite{tarvainen2017mean}, where the teacher model $f_{{\theta}^{'}}$ uses the exponential
moving average (EMA) weights of the student model $f_{{\theta}}$, \ie,
$\theta^{'}_{t} = \alpha \theta^{'}_{t-1} + (1-\alpha) \theta_{t}$. 
\reviselast{Specifically, the weight of teacher model $\theta_{t}^{'}$  is updated through $\theta_{t}^{'} = \alpha \theta^{'}_{t-1} + ( 1 - \alpha) \theta_{t} $, where $\theta_{t}$ is the student model parameters and $\theta_{t}^{'}$ is the teacher model parameters. $\alpha$ is a smoothing coefficient hyperparameter that affects how the teacher model relies on the current student model parameter. If $\alpha$ is large, \reviseagain{the teacher model relies more} on the previous teacher model in the last step; otherwise, \reviseagain{the teacher model relies more} on the current student model parameters. According to the empirical evidence in~\cite{tarvainen2017mean}, setting $ \alpha $ = 0 makes the model as a variation of the $\pi$ model, and the performance is the best when setting $\alpha$ =  0.999. We follow this empirical experience and set $\alpha$ to 0.999 in our experiments.}
Then, the transformation-consistent regularization is performed on the input to the teacher model $f_{{\theta}^{'}}$, and the consistency loss is added on the two predictions of the teacher and student model, respectively.

\subsection{Transformation Consistent Self-ensembling Model}

Next, we introduce how we design the randomized data transformation regularization for segmentation,~\ie, the transformation-consistent self-ensembling model (TCSM\_v2).

\subsubsection{Motivation}
\reviseagain{
In general self-ensembling semi-supervised learning, most regularization and perturbations can be easily designed for the classification problem.
However, in the medical image domain, accurate segmentation of important structures or lesions is a very challenging problem and the perturbations for segmentation tasks are more worthy to explore.
One prominent difference between these two common tasks is that the classification problem is transformation \emph{invariant} while the segmentation task is expected to be transformation \emph{equivariant}.}
Specifically, for image classification, the convolutional neural network only recognizes the presence or absence of an object in the whole image. In other words, the classification result should remain the same, no matter what the data transformation (\ie, translation, rotation, and flipping) are applied to the input image. 
\reviseagain{While in the image segmentation task,} if the input image is rotated, the segmentation mask is expected to have the same rotation with the original mask, although the corresponding pixel-wise predictions are the same; see examples in Figure~\ref{fig:method2}~(a).
However, in general, convolutions are not transformation (\ie, flipping, rotation) equivariant\footnote{Transformation in this work refers to flipping, scaling and rotation.}, meaning that if one rotates or flips the CNN input, then the feature maps do not necessarily rotate in a meaningful manner~\cite{worrall2017harmonic}, as shown in Figure~\ref{fig:method2}~(b).
Therefore, the convolutional network consisting of a series of convolutions is also not transformation equivariant. 
Formally, every transformation $ \pi \in \Pi$ of input $\textbf{x}$ associates with a transformation $\psi \in \Psi$  of the outputs; that is
$
\psi[f(\textbf{x})] = f(\pi[\textbf{x}]),
$
but in general $\pi \neq \psi$.

\subsubsection{\reviseagain{Mechanism of TCSM}}
\reviseagain{This phenomenon limits the unsupervised regularization effect of randomized data transformation for segmentation~\cite{laine2016temporal}.}
To enhance the regularization and more effectively utilize unlabeled data in our segmentation task, we introduce a transformation-consistent scheme in the unsupervised regularization term. 
Specifically, this transformation-consistent scheme is embedded in the framework by approximating $\psi$ to $\pi$ at the input and output space.
Figure~\ref{fig:pipeline} shows the detailed illustration of the framework and Algorithm~\ref{alg1} shows the pseudocode.
Under the transformation-consistent scheme and other different perturbations (\eg, Gaussian noise and network dropout), each input $x_i$ is fed into the network for twice evaluation to acquire two outputs $z_i$ and $\tilde{z}_i$.
More specifically, the transformation-consistent scheme consists of triple $\pi_i$ operations; see Figure~\ref{fig:pipeline}. 
For one training input $x_i$, in the first evaluation, the operation $\pi_i$ is applied to the input image while in the second evaluation, the operation $\pi_i$ is applied to the prediction map. 
Random perturbations (\eg, Gaussian noise and network dropout) are applied in the network during the twice evaluations. 
By minimizing the difference between $z_i$ and $\tilde{z}_i$ with a mean square error loss, the network is regularized to be transformation-consistent and thus increase the network generalization capacity. 
Notably, the regularization loss is evaluated on both the labeled and unlabeled inputs.
To utilize the labeled data $x_i \in \mathcal{L}$, the same operation $\pi_i$ is also performed on $y_i$ and optimized by the standard cross-entropy loss. 
Finally, the network is trained by minimizing a weighted combination of unsupervised regularization loss and supervised cross-entropy loss.

\reviseagain{
\begin{algorithm}[tb]  
	\caption{\revise{TCSM\_v2 Algorithm pseudocode.}  }
	\label{alg:Framwork} 
	\begin{algorithmic}[1]  
		\Require $x_i \in {\mathcal{L}+\mathcal{U}} , y_i\in \mathcal{L}$
		\revise{		\State $\lambda(T)$ =  unsupervised weight function
			\State
			$f_{\theta}(x)$ = student model with parameters $\theta$
			\State
			$f_{\theta^{'}}(x)$ = teacher model  with parameters $\theta^{'}$
			\State
			$\pi_i(x)$ =  transformation operations
			\State 
			$\alpha$ = smoothing coefficient hyperparameter. 
			\For {$T$ in $[1, numepochs]$}
			\For {each minibatch $B$}
			\State randomly update $\pi_i(x)$
			\State  $z_{i \in B}$ $\leftarrow \pi_i(f_{\theta}(x_{i\in B})) $ 
			
			\State    	$ \tilde{z}_{i \in B} \leftarrow f_{\theta^{'}}(\pi_i(x_{i\in B})) $ 
			
		\reviseagain{	\State $loss \leftarrow - \frac{1}{\left | B^{'} \right |}\sum_{i\in(B^{'})} {\rm log} z_i[\pi_i(y_i)]  + $ 
			\State  $\lambda(T)\frac{1}{\left | B \right |}\sum_{i\in B} 
			\left \| z_i - \tilde{z}_i \right \|^ 2$ }
			\State  update $\theta$ using optimizer
		\reviseagain{	\State  update $ \theta^{'}_{T} \leftarrow  \alpha \theta^{'}_{T-1} + (1-\alpha) \theta_{T}$ }
			\EndFor {}
			\EndFor }
		\State	
		\Return $\theta^{'}$; 
	\end{algorithmic}  
	\label{alg1}
\end{algorithm} 

\subsubsection{Loss function}
It has cross-entropy loss on the labeled inputs and the regularization term on both the labeled and unlabeled inputs. The overall loss function is then defined as 

\vspace*{-2.5mm}
\begin{equation}
loss = \mathcal{L} + \lambda(T) \mathcal{R} ,
\vspace*{-0.5mm}
\end{equation}
where $\mathcal{L}$ and $\mathcal{R}$ are the supervised term and regularization term, respectively. 
The time-dependent warming up function $\lambda(T)$ is a weighting factor for supervised loss and regularization loss.
This weighting function is a Gaussian ramp-up curve, \ie, $\lambda(T) = k*e^{(-5(1-T)^2)}$, where $T$ denotes the training epoch and $k$ scales the maximum value of the weighting function. In our experiments, we empirically set $k$ as 1.0.

We randomly sample $x_{i \in B}$ images from the training data and the supervised term $\mathcal{L}$ within one minibatch is defined as 

\begin{equation}
\vspace*{-2.5mm}
\mathcal{L} = - \frac{1}{\left | B^{'} \right |}\sum_{i\in B^{'}} {\rm log} z_i[\pi_i(y_i)],
\vspace*{-0.3mm}
\end{equation}
where $B^{'} \in B$ denotes the labeled images within a minibatch, $z_i$ and $y_i$ are the network prediction and ground-truth segmentation label, respectively. 
At the beginning of the network training, $\lambda (T)$ is small and training is mainly dominated by the supervised loss on the labeled data. In this way, the network is able to \reviseagain{learn accurate information} from the labeled data. As the training progresses, the network gets a reliable model and can generate output for the unlabeled data. 
The regularization term optimizes the prediction differences by calculating the differences on the predictions.
\begin{equation}
\mathcal{R} = \frac{1}{\left | B \right |}\sum_{i\in B} 
\left \| z_i - \tilde{z}_i \right \|^ 2,
\end{equation}
where $z_i$ and $\tilde{z}_i$ denote the network predictions of the student model and teacher model, respectively, \ie,  $z_i = \pi_i (f_\theta (x_{i \in B}) )$ and  $\tilde{z}_i =  f_{\theta^{'}} (\pi_i ( x_{i \in B} ))$. $f_{\theta}$ and $f_{\theta^{'}}$ denote the student model and teacher model, respectively. The student model $f_\theta$ is updated by the gradient descent while the teacher model is updated by ${\theta^{'}_{T}} = \alpha {\theta^{'}_{T-1}} + (1-\alpha) \theta_T$, where $\alpha = 0.999$.  


\subsubsection{Implementation of $\pi$ operation}
Multi-scale ensembling is shown to be effective for image recognition~\cite{he2016deep,he2015spatial,jegou2010product}.
To enlarge the regularization effect for semi-supervised learning, we extend the TCSM to a more generalized form including random scaling. 
The main goal is to keep the consistency of the teacher and student model after multi-scale inference. 
Specifically, $\pi_i$ operation includes not only rotation but also random scaling operation. 
For the student model, we give the input and generate the prediction $z_i$. 
For the teacher model, we randomly scale the input image and generate the prediction result, which is then rescaled to the original size of the input image and we finally get $\tilde{z_i}$.
Then, two predictions $z_i$ and  $\tilde{z_i}$ are minimized through a mean square error loss. 
}

The transformation-consistent scheme includes random scaling, the horizontal flipping operation as well as four kinds of rotation operations to the input with angles of $\gamma \cdot 90^{\circ}, $ where $ \gamma\in \left \{0,1,2,3 \right \}$.
During each training pass, one rotation operation and one scaling operation within the scaling ratio of 0.8-1.2 is randomly chosen and applied to the input image, \ie, $\pi_i (x_i)$.
To keep two terms in the loss function, we evenly and randomly select the labeled and the unlabeled samples in each minibatch.
Note that we employed the same data augmentation in the training procedure of all the experiments for a fair comparison. 
However, our method is different from traditional data augmentation.
Specifically, our method utilized the unlabeled data by minimizing the network output difference under the transformed inputs, while complying with the \emph{smoothness} assumption.

\subsection{\reviseagain{Technical Details of TCSM\_v2}}

\begin{figure}[!t]
	\centering
	\includegraphics[width=0.99\linewidth]{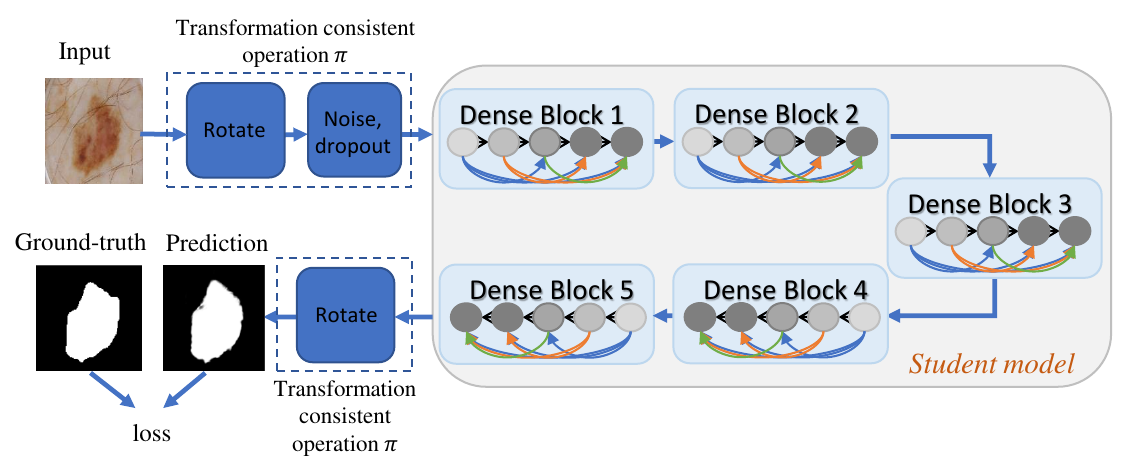}
	\vspace*{-4mm}
	\caption{\reviselast{The visualization of transformation-consistent operation $\pi$ in the DenseUnet architecture. We omit the unet connection and the decoder part for simplification.}}
	\label{fig:tcsm_figure}\centering
	\vspace*{-3mm}
\end{figure}

For dermoscopy images and retinal fundus images, we employ the 2D DenseUNet architecture~\cite{li2018h} as \reviseagain{both our teacher and student models}.
Compared to the standard DenseNet~\cite{huang2017densely}, we add the decoder part for the segmentation tasks. The decoder has four blocks and each block consists of "upsampling, convolutional, batch normalization and ReLU activation" layers. 
The UNet-like skip connection is added between the final convolution layer of each dense block in the encoder part and the convolution layer in the decoder part. 
The final prediction layer is a convolutional layer \reviseagain{with a channel number} of two.
Before the final convolution layer, we add a dropout layer with a drop rate of 0.3.
\reviseagain{The network was trained with Adam algorithm~\cite{kingma2014adam} with a learning rate of 0.0001. All the experiments are trained for a total of 8000 iterations.}
We also visualize the network structure diagram to illustrate the implementation of TCSM in Figure~\ref{fig:tcsm_figure}.

To generalize our method to 3D medical images, \eg, liver CT scans, we train TCSM\_v2 with  3D U-Net~\cite{cciccek20163d}.
For training with 3D U-Net, we follow the original setting with the following modifications. We modify the base filter parameters to 32 to accommodate this input size. 
The optimizer is SGD with a learning rate of 0.01. The batch normalization layer is employed to facilitate the training process and the loss function is modified to the standard weighted cross-entropy loss. 
All the experiments are trained for \reviseagain{a total of 9000 iterations}.

\reviseagain{
We implemented the model using PyTorch~\cite{paszke2017automatic}.
The experiments differ slightly from that in~\cite{li2018semi} due to the different implementation platforms.
We used the standard data augmentation techniques on-the-fly to avoid overfitting, including randomly flipping, rotating, and scaling with a random scale factor from 0.9 to 1.1.
Note that all the experiments employed data augmentation for a fair comparison.
In the inference phase, we remove the transformation operations in the network and do one single test with the original input for a fair comparison. 
After getting the probability map from the network, we first apply thresholding with 0.5 to generate the binary segmentation result, and then use morphology operation, \ie, filling holes, to obtain the final segmentation result.}

\section{Experiments}
\subsection{Datasets}
To evaluate our method, we conduct experiments on various modalities of medical images, including dermoscopy images, retinal fundus images, and liver CT scans.

\para{Dermoscopy image dataset.} 
The dermoscopy image dataset in our experiments is the 2017 ISIC skin lesion segmentation challenge dataset~\cite{codella2018skin}.
It includes a training set with 2000 annotated dermoscopic images, a validation set with 150 images, and a testing set with 600 images.
The image size ranges from $540 \times 722$ to $4499 \times 6748$. 
\reviseagain{To balance the segmentation} performance and computational cost, we first resize all the images to $248 \times 248$ using bicubic interpolation.

\para{Retinal fundus image dataset.} The fundus image dataset is~\reviseagain{from the} MICCAI 2018 Retinal Fundus Glaucoma Challenge (REFUGE)\footnote{https://refuge.grand-challenge.org/REFUGE2018/}.
Manual pixel-wise annotations of the optic disc were obtained by seven independent ophthalmologists from the Zhongshan Ophthalmic Center, Sun Yat-sen University. 
\reviseagain{Experiments were conducted on the released training dataset, which contains 400 retinal images. }
The training dataset is randomly split to training and test sets, and we resize all the images to $248 \times 248$ using bicubic interpolation.

\reviselast{
\para{Liver segmentation dataset.} The liver segmentation dataset are from the 2017 Liver Tumor Segmentation Challenge (LiTS)\footnote{https://competitions.codalab.org/competitions/17094\#participate-get\_data}~\cite{bilic2019liver,seo2019modified}.}
The LiTS dataset contains 131 and 70 contrast-enhanced 3D abdominal CT scans for training and testing, respectively.
The dataset was acquired by different scanners and protocols at six different clinical sites, with a largely varying in-plane resolution from 0.55 mm to 1.0 mm and slice spacing from 0.45 mm to 6.0 mm.

\subsection{Evaluation Metrics}
For dermoscopy image dataset, we use \reviseagain{Jaccard index} (JA), dice coefficient (DI), pixel-wise accuracy (AC), sensitivity (SE) and specificity (SP) to measure the segmentation performance: 
\vspace*{-3mm}
\begin{gather}
AC =\frac{TP+TN}{TP+FP+TN+FN}, \notag \\
SE = \frac{TP}{TP+FN}, \quad SP = \frac{TN}{TN+FP},\\
JA = \frac{TP}{TP+FN+FP}, \quad DI =\frac{2\cdot TP}{2 \cdot TP + FN + FP}, \notag
\end{gather}
where $TP, TN, FP $, and $FN$ refer to the number of true positives, true negatives, false positives, and false negatives, respectively.  
\reviseagain{For the retinal fundus image dataset}, we use JA to measure the optic disc segmentation accuracy. 
\reviseagain{For the liver CT dataset}, Dice per case score is employed to measure the accuracy of the liver segmentation result, according to the evaluation of \reviseagain{the 2017 LiTS challenge}~\cite{bilic2019liver}.

\subsection{Experiments on Dermoscopy Image Dataset}
\begin{table}[!t]
	\centering
	\caption{\revise{Comparison of supervised learning and semi-supervised learning (50 labeled/1950 unlabeled) on the validation set in the dermoscopy image dataset. "Supervised+reg" denotes supervised with regularization.}}
	\vspace*{-1mm}
	{\resizebox{0.8\columnwidth}{!}{	\begin{tabular}{cccc}
				\toprule[1.5pt] 
				Metric & Supervised & Supervised+regu & Ours  \tabularnewline \hline
				
				JA & 71.17 & 72.28  & \textbf{75.24}  \tabularnewline 
				
				DI & 79.91 & 81.10 & \textbf{83.44}  \tabularnewline 
				
				AC & 91.95 & 93.52 & \textbf{94.46}  \tabularnewline

				SE & 75.90 & 81.17 & \textbf{83.07}  \tabularnewline 
				
				SP & 97.04 & 97.02 & \textbf{97.07}   \tabularnewline 
				
				\bottomrule[1.5pt]	 
			\end{tabular}}
			\label{tab:ablation}}
	\vspace*{-3mm}
\end{table}

\begin{figure}[!t]
	\centering
	\includegraphics[width=0.8\linewidth]{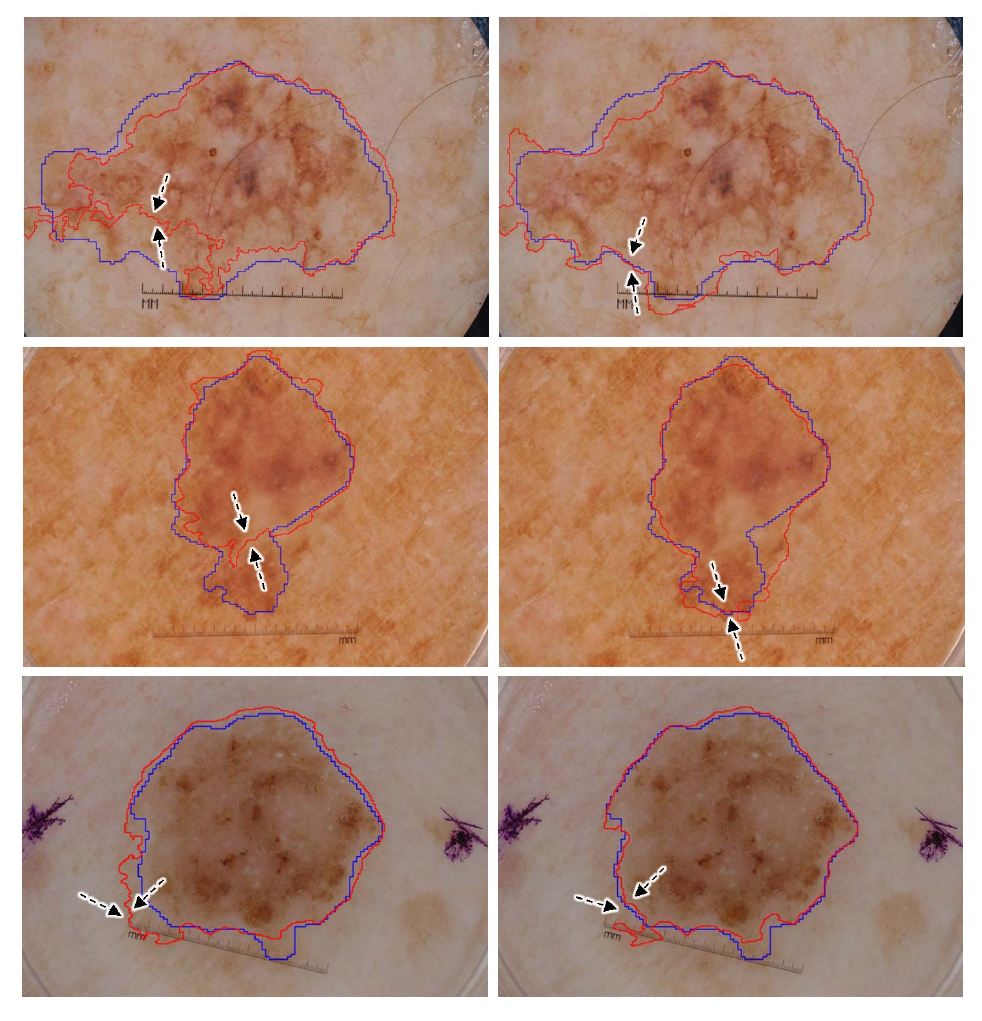}
	\vspace*{-1mm}
	\caption{
		Examples of the segmentation results of supervised learning (left) and our method (right) on the validation set in the dermoscopy image dataset. The blue and red contours denote the ground truth and our segmentation result, respectively.}
	\label{fig:result}\centering
	\vspace*{-3mm}
\end{figure}

\subsubsection{Quantitative and visual results with 50 labeled data}
We report the performance of our method trained with only 50 labeled images and 1950 unlabeled images.
Note that the labeled image is randomly selected from the whole dataset.
Table~\ref{tab:ablation} shows the experiments with \reviseagain{the supervised method}, supervised with regularization, and our semi-supervised method on the validation dataset. 
We use the same network architecture (DenseUNet) in all these experiments \reviseagain{for a fair comparison}.
The supervised experiment is optimized by the standard cross-entropy loss on the 50 labeled images.
The supervised with regularization experiment is also trained with 50 labeled images, but differently, the total loss function is \reviseagain{a weighted combination} of the cross-entropy loss and the regularization loss, which is the same with our  loss function. 
Our method is trained with 50 labeled and 1950 unlabeled images \reviseagain{in a semi-supervised manner}.
\revise{
From Table~\ref{tab:ablation}, it is observed that our semi-supervised method achieves higher performance than a supervised counterpart on all the evaluation metrics, with prominent improvements of 4.07\% on JA and 3.47\% on DI, respectively. 
It is worth mentioning that supervised with regularization experiment improves the supervised training due to the regularization loss on the labeled images; see "Supervised+regu" in Table~\ref{tab:ablation}. The consistent improvements of "Supervised+regu" on all evaluation metrics demonstrate the regularization loss is also effective for the labeled images. 
Figure~\ref{fig:result} presents some segmentation results (red contour) of supervised method (left) and our method (right).}
Comparing with the segmentation contour achieved by the supervised method (left column), the semi-supervised method fits more consistently with the ground-truth boundary.
The observation shows the effectiveness of our semi-supervised learning method, compared with the supervised method.

%
%
%
%
%
%
%
%
%

\begin{table}[t]
	\centering
	\caption{\reviseagain{Ablation of semi-supervised method (50 labeled/1950 unlabeled) on the validation set in dermoscopy image dataset. ``R'' denotes rotation transformation. ``ND'' denotes Gaussian noise and dropout. ``NDR'' denotes Gaussian noise, dropout, and rotation. ``Shift ($r$)'' denotes shifting transformation with ratio $r$.  ``NDRscale'' denotes the transformations including noise, dropout, rotation, and random scaling. (Unit: \%).}}
	{
{\resizebox{1.0\columnwidth}{!}{	\begin{tabular}{c|ccccc}	
\toprule[1.5pt] 

Setting & JA & DI & AC & SE & SP  \tabularnewline \hline

Supervised & 71.17 & 79.91  & 91.95  & 75.90 &  97.04 \tabularnewline 

TCSM-R & 73.76 &   82.05& 94.87 & 84.14 & \textbf{97.65} \tabularnewline 
TCSM-ND & 74.03 &   82.33 &  94.30 & 83.84 & 96.77 \tabularnewline

TCSM-NDR  & 74.46 &  82.22 &  \textbf{95.06} & 82.51 & 97.42 \tabularnewline

\hline 

TCSM\_v2-NDR & 74.91
& 82.90 & 94.63 & 82.80 & 97.61 \tabularnewline
\reviseagain{
TCSM\_v2-Shift(0.3)} & 71.26 &  75.11 & 79.5   & 78.32 & 85.40  \tabularnewline 
\reviseagain{
	TCSM\_v2-Shift(0.2)} & 71.69 &  79.24 & 92.34 & 80.12 & 95.42 \tabularnewline
\reviseagain{
TCSM\_v2-Shift(0.1)} & 72.74 &  80.93 & 93.35 & 81.32 & 96.62 \tabularnewline
\reviseagain{ 
TCSM\_v2-Scale(0.3)}  & 71.62 & 79.96 & 92.69  & 79.86 & 97.76 \tabularnewline
\reviseagain{
	TCSM\_v2-Scale(0.2)} & 71.87 &  79.78 & 92.75 & 79.98 & 96.04 \tabularnewline
\reviseagain{ 
TCSM\_v2-Scale(0.1)} & 73.57 & 81.51 & 94.24 & 80.57 & 96.90 \tabularnewline 

\textbf{ TCSM\_v2-NDRScale(\textbf{ours})}  & \textbf{75.24} 
& \textbf{83.44} & 94.46 & \textbf{83.07} & 97.07 \tabularnewline

\bottomrule[1.5pt]
\end{tabular}}}}
\label{tab:abla}
\vspace*{-3mm}
\end{table}

\subsubsection{Effectiveness of TCSM and TCSM\_v2}
\revise{
To show the effectiveness of our proposed transformation-consistent method, we conducted an ablation analysis of our method on the dermoscopy image dataset, as the results are shown in Table~\ref{tab:abla}. 
The experiments were performed with randomly-selected 50 labeled data and 1950 unlabeled data, and tested on the validation set. 
In the ``Supervised'' setting, we trained the network with only 50 labeled data. 
``TCSM-ND'' refers to semi-supervised learning with Gaussian noise and dropout regularization. ``TCSM-R'' refers to semi-supervised learning with transformation-consistent regularization (only rotation), and ``TCSM'' refers to the experiment with all of these regularizations.
As shown in Table~\ref{tab:abla}, both kinds of regularizations independently contribute to the performance gains of semi-supervised learning.
The resulting improvement with transformation-consistent regularization is very competitive, compared with the performance increment with Gaussian noise and dropout regularizations.
\reviseagain{These two regularizations are complementary, so when they are employed together, the performance can be further enhanced.}

\reviseagain{
Moreover, we analyze the effects of the shifting and scaling operations. Shift ($r$) denotes randomly shifting the image by $r'W$ or $r'H$, where $ r' \in [1-r, 1+r] $, and $W $ and $H$ denote the image width and height. Scale ($r$) denotes randomly scaling the image to $(r'W,r'H)$, where $r' \in [1-r , 1+r]$. From the experiments in Table~\ref{tab:abla}, we can see that random scaling with ratio 0.1 could improve the semi-supervised learning results, while other transformation settings have limited improvements. }
``TCSM\_v2-NDR'' denotes the mean teacher based semi-supervised learning with transformation-consistent strategy (only rotation). 
``TCSM\_v2-NDRScale'' refers to the mean teacher based semi-supervised learning with our transformation-consistent strategy including both rotation and scaling. 
From these two comparisons, we can see that the generalized form of transformation-consistent strategy improves the semi-supervised learning. 
``TCSM'' and ``TCSM\_v2-NDR'' utilizes the same regularization. From these two comparisons, we can find that the weight-averaged consistency targets improve the semi-supervised deep learning results. 
Our final model achieves 75.24\% JA and 83.44\% DI, surpassing the supervised baseline by 5.7\% JA and 4.4\%  DI. 
}

\subsubsection{Results under different number of labeled data}
\begin{table}[t]
	\centering
	\caption{\revise{Results of our method on the validation set under different number of labeled/unlabeled images (Unit:\%).}}
	{\resizebox{0.95\columnwidth}{!}{
	\begin{tabular}{c|c|ccc}
		\toprule[1.5pt] 
		Label/Unlabel & Metric & Supervised  & TCSM & TCSM\_v2 \tabularnewline \hline
		
		\multirow{5}{*}{50/1950} & JA & 71.17  & 74.46 & \textbf{75.24} \tabularnewline 
		
		& DI & 79.91 & 82.22 & \textbf{83.44} \tabularnewline 
		
		& AC & 91.95 & \textbf{95.06}  & 94.46 \tabularnewline

		& SE & 75.90 & 82.51  & \textbf{83.07} \tabularnewline 
		
		& SP & 97.04  & \textbf{97.42}  & 97.07 \tabularnewline 
		\hline
		\multirow{5}{*}{100/1900} & JA & 73.92  & 74.75 & \textbf{75.52}  \tabularnewline 
		
		& DI & 82.37 & 83.02 &  \textbf{83.35}  \tabularnewline 
		
		& AC & 93.87 &  \textbf{94.03} & 93.28  \tabularnewline

		& SE & 83.95 & \textbf{86.02} & 83.78 \tabularnewline 
		
		& SP & \textbf{97.08} & 96.89 & 94.65 \tabularnewline
		\hline 
		\multirow{5}{*}{300/1700} & JA & 76.66 & 77.19 & \textbf{77.52} \tabularnewline 
		
		& DI & 84.32 & \textbf{85.38} & 85.20 \tabularnewline 
	
		& AC & 94.25  & 95.11 & \textbf{95.90} \tabularnewline 
			
		& SE & 85.82  & 86.02 & \textbf{86.31} \tabularnewline 
		
		& SP & 95.77  & 95.78 &  \textbf{95.85}  \tabularnewline 
		\hline
		\multirow{5}{*}{2000/0} & JA &  78.80 & 79.15 & \textbf{79.27} \tabularnewline 
		
		& DI & 86.67 & 87.01 & \textbf{88.13} \tabularnewline 
		
		& AC & 95.12 &  95.03 & \textbf{96.01} \tabularnewline 
		
		& SE & 88.75 & 89.20 & \textbf{89.35} \tabularnewline 
		
		& SP &  97.02 & 96.91 & \textbf{97.14} \tabularnewline 
		\bottomrule[1.5pt]	 
	\end{tabular}}}
	\label{tab:port}
\end{table}

\begin{figure}[!t]
	\centering
	\includegraphics[width=0.65\linewidth]{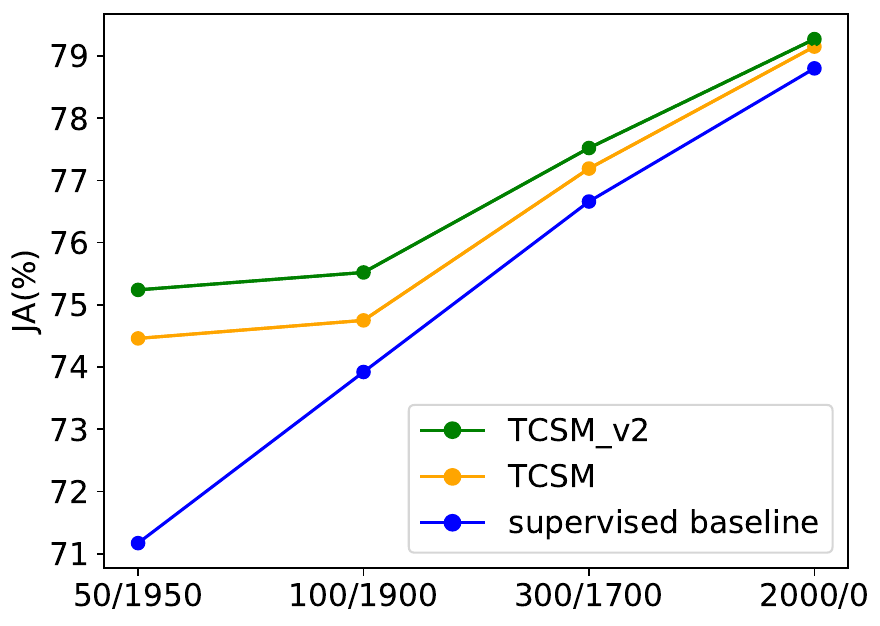}
	\vspace*{-1mm}
	\caption{\revise{Results on the validation set of the dermoscopy image dataset with different number of labeled/unlabeled data.}}
	\label{fig:difnumber}\centering
\end{figure}

\revise{

Table~\ref{tab:port} shows the lesion segmentation results of our TCSM and TCSM\_v2 (trained with labeled data and unlabeled data) and supervised method (trained only with labeled data) under a different number of labeled/unlabeled images.
We draw the JA score of the results in Figure~\ref{fig:difnumber}.
It is observed that the semi-supervised methods consistently performs better than the supervised method in different labeled/unlabeled data settings, \reviseagain{demonstrating that our method effectively utilizes the unlabeled data and brings} performance gains.
Note that in all semi-supervised learning experiments, we train the network with 2000 images in total, including labeled images and unlabeled images. }
As expected, the performance of supervised training increases when more labeled training images are available; see the blue line in Figure~\ref{fig:difnumber}.
\reviselast{
At the same time, the segmentation performance of semi-supervised learning can also \reviseagain{increase} with more labeled training images; see the orange line in Figure~\ref{fig:difnumber}.
The performance gap between supervised training and semi-supervised learning narrows as more labeled samples are available, which conforms with our expectations.}
When the amount of labeled dataset is small, our method can gain a large improvement, since the regularization loss can effectively leverage more information from the unlabeled data.
Comparatively, as the number of labeled data increases, the improvement becomes limited. 
This is because the labeled and unlabeled data are randomly selected from the same dataset and a large amount of labeled data may reach the upper bound performance of the dataset.  

\revise{
From the comparison between TCSM and TCSM\_v2, we can see that TCSM\_v2 consistently improve TCSM under \reviseagain{different label and unlabeled settings}.
From the comparison between the semi-supervised method and supervised method trained with 2000 labeled images in Figure~\ref{fig:difnumber}, it can be observed that our method increases the JA performance when all labels are used.
The improvement indicates that the unsupervised loss can also provide a regularization to the labeled data.
In other words, the consistency requirement in the regularization term can encourage the network to learn more robust features to improve the segmentation performance.
}

\subsubsection{Comparison with other semi-supervised segmentation methods}

\reviselast{			
We compare our method with two latest semi-supervised segmentation methods~\cite{bai2017semi,zhang2017deep} in the medical imaging community and an adversarial learning-based semi-supervised method~\cite{Hung_semiseg_2018}.
Also, we extend the semi-supervised classification model~\cite{tarvainen2017mean} to segmentation for comparison. 
Note that the method~\cite{nie2018asdnet} for medical image segmentation adopts a similar idea with the adversarial learning-based method~\cite{Hung_semiseg_2018}. }

\begin{table}[t]
	\centering
	\caption{\revise{JA performance of different semi-supervised methods on the validation dataset \reviseagain{of the dermoscopy image dataset.} 
		``Supervised'' denotes training with 50 labeled data (Unit: \%).}}
	\vspace*{-1mm}
	\label{tab: semi_comparision} %
	{\resizebox{1.0\columnwidth}{!}	{
		\begin{tabular}{c|c|c|c}
			\toprule[1.5pt]
			Method 		& Backbone			& Result & Improvement \tabularnewline \hline 
			
			Supervised & \multirow{6}{*}{DenseUNet} & 71.17 & -
			\tabularnewline
			
			GAN~\cite{Hung_semiseg_2018} &  & 73.31 & 2.14 \tabularnewline 
			DAN~\cite{zhang2017deep} &  & 73.82 & 2.65\tabularnewline

			Mean Teacher~\cite{tarvainen2017mean} 	& 	& 74.23	& 3.06 \tabularnewline
			Self-training~\cite{bai2017semi}	 &   & 74.40  & 3.23 \tabularnewline 
			
			Ours	& 	& \textbf{75.24}		& \textbf{4.07} \tabularnewline
			\bottomrule[1.5pt]	     
		\end{tabular}}}
		\vspace*{-3mm}
\end{table}

\revise{
For a fair comparison, we~\reviseagain{re-implemented} their methods with the same network backbone on this dataset. All the experiments utilized the same data augmentation and training strategies.
We~\reviseagain{conducted experiments} with the setting of 50 labeled images and 1950 unlabeled images.
Table~\ref{tab: semi_comparision} shows the JA performance of different methods on the validation set.
As shown in Table~\ref{tab: semi_comparision}, our method achieves 4.07\% JA improvement by utilizing unlabeled data. 
Compared with other methods, we achieve the highest improvement over the supervised baseline. 
The comparison shows the effectiveness of our semi-supervised segmentation method, compared to other semi-supervised methods.
}

\subsubsection{Comparison with methods on the challenge leaderboard}

\revise{
We also compare our method with state-of-the-art methods submitted to the ISIC 2017 skin lesion segmentation \reviseagain{challenge~\cite{codella2018skin}}.
\reviseagain{There are a total of 21 submissions} and the top results are listed in Table~\ref{tab: testdata}.
Note that the final rank is determined according to JA on the testing set.
We trained two models: TCSM\_v2 and baseline. TCSM\_v2 was trained with 300 labeled images and the left are utilized as the unlabeled images. The baseline model is trained with only 300 labeled data. 
Other methods use all labeled data as the training data.
The supervised model is denoted as our baseline model. 
As shown in Table~\ref{tab: testdata}, our semi-supervised method achieved the best performance on the benchmark, outperforming the state-of-the-art method~\cite{yuan2017improving} with 1.6\% improvement on JA (from 76.5\% to 78.1\%). 
The performance gains on DI and SE are consistent with that on JA, with 1.1\% and 3.7\% improvement, respectively.
\reviseagain{Our baseline model with 300 labeled data also outperforms} other methods due to state-of-the-art network architecture.
Based on this strong baseline, our semi-supervised learning method further makes significant improvements, which demonstrates the effectiveness of the overall semi-supervised learning method.
}

\begin{table}[!t]
	\centering
	\caption{\revise{Results on the test dataset in the ISIC 2017 dermoscopy lesion segmentation challenge (Unit: \%).}}
	\vspace*{-1mm}
	\label{tab: testdata}
	{	{
			\begin{tabular}{c|c|c|c|c|c|c}
				\toprule[1.5pt]
				Team  & Labels  & JA  & DI  & AC & SE & SP  \tabularnewline  \hline 
		
%
				
				Our semi-supervised& \multirow{2}{*}{300}
				& \textbf{78.1} & \textbf{86.0}  & \textbf{94.1}  & \textbf{86.2} &  96.8 
				\tabularnewline
				
				Our baseline  &
				& 76.8 &   84.8 & 93.6  & 83.6 & 96.7   \tabularnewline
				\hline

				\citet{yuan2017improving} &  \multirow{9}{*}{2000} & 76.5 & 84.9   & 93.4  & 82.5 & 97.5   \tabularnewline
				
				\citet{venkatesh2018deep}& 	 & 76.4 & 85.6 & 93.6 & 83.0 & 97.6 \tabularnewline

				\citet{berseth2017isic}	&  & 76.2 & 84.7 &  93.2 & 82.0 & 97.8  \tabularnewline
				
				\citet{bi2017automatic}&    & 76.0 & 84.4 &  93.4  & 80.2 & \textbf{98.5}  \tabularnewline
				
				RECOD	 &  & 75.4 & 83.9 & 93.1 & 81.7 & 97.0  \tabularnewline
				
				Jer	&  & 75.2 & 83.7 & 93.0 & 81.3 & 97.6  \tabularnewline 
				
				NedMos	 &  & 74.9 & 83.9 &  93.0 & 81.0 & 98.1  
				\tabularnewline
				
				INESC	 &  & 73.5 & 82.4 & 92.2 & 81.3 & 96.8  
				\tabularnewline
				
				Shenzhen U (Lee)	&   & 71.8 & 81.0 &  92.2 & 78.9  & 97.5  
				\tabularnewline
				
				
				\bottomrule[1.5pt]	    
		\end{tabular}}
	}
	\vspace*{-1mm}
\end{table}

\begin{table}[!t]
	\centering
	\caption{\reviselast{JA performance of different methods on the fundus image dataset. ``10\%'' and ``20\%'' denote training with ``10\%'' and ``20\%'' labeled data in the training set, respectively. ``Imp'' refers to the improvement over the supervised baseline.}}
	\label{tab:retina} %
	{\resizebox{1.0\columnwidth}{!}{
			\begin{tabular}{c|c|c|c|c|c}
				\toprule[1.5pt]
				Method 		& Backbone			& 10\%  & Imp &  20\%  & Imp \tabularnewline \hline 
				
				Supervised &  \multirow{6}{*}{DenseUNet} & 93.61 & - & 94.61 & -
				\tabularnewline 
				
				Self-training~\cite{bai2017semi}	 &   & 94.20  & 0.59 & 95.01 &  0.40 \tabularnewline 
				
				GAN~\cite{Hung_semiseg_2018} & 	 & 94.32 & 0.71 & 94.93 & 0.32 \tabularnewline 
				DAN~\cite{zhang2017deep}	 &   & 94.47  & 0.86  & 94.87 &  0.26 \tabularnewline 
				
				Mean Teacher~\cite{tarvainen2017mean}	 &   & 94.76 & 1.15  &  95.02 & 0.41 \tabularnewline 
				\textbf{Ours}  	& 		& \textbf{95.43} 	& \textbf{1.82} & \textbf{95.35} & \textbf{0.74} \tabularnewline
				\bottomrule[1.5pt]	     
			\end{tabular}}}
\end{table}

\begin{table}[!t]
	\centering
	\caption{\reviselast{Dice performance of different semi-supervised methods on the LiTS dataset. ``10\%'' and ``20\%'' denote training with ``10\%'' and ``20\%''labeled data, respectively. ``Imp'' refers to the improvement over the supervised baseline.}}
	\label{tab:lits} %
	\vspace*{-1mm}
	{\resizebox{1.0\columnwidth}{!}	{\begin{tabular}{c|c|c|c|c|c}
				\toprule[1.5pt]
				Method & Backbone & 10\%  & Imp & 20\%  &  Imp
				\tabularnewline \hline 
				Supervised & \multirow{6}{*}{3D U-Net}  & 87.97 & - & 88.55 & - 	\tabularnewline
				Self-training~\cite{bai2017semi} &  & 89.64 & 1.67 & 89.65 & 1.10	\tabularnewline
				DAN~\cite{zhang2017deep} &   & 92.90 & 4.93 & 92.29 & 3.74
				\tabularnewline
				GAN~\cite{Hung_semiseg_2018} &   & 92.93 & 4.96 & 93.75 & 5.20 \tabularnewline
				Mean Teacher~\cite{tarvainen2017mean} &   &  93.28 &  5.31 &  93.37 & 4.82  
				\tabularnewline
				\textbf{Ours} &  & \textbf{93.30} & \textbf{5.33} & \textbf{94.27} & \textbf{5.72}				\tabularnewline
				\bottomrule[1.5pt]	  
	\end{tabular}}
}
\vspace*{-3mm}
\end{table}

\subsection{Experiments on Retinal Fundus Image Dataset}

We report the performance of our method for optic disc segmentation from retinal fundus images. 
The 400 training images from~\reviseagain{the REFUGE challenge~\cite{orlando2020refuge}} were randomly separated to training and test dataset with the ratio of 9:1.
For training semi-supervised model, only a portion of labels (\ie, 10\% and 20\%) in the training set were used. 
We preprocessed all the input images by subtracting the mean RGB values of all the training dataset. 
When training the supervised model, the loss function was the traditional cross-entropy loss and \reviseagain{we used the SGD algorithm} with learning rate 0.01 and momentum 0.9. 
To train the semi-supervised model, we added the extra unsupervised regularization loss, and the learning rate was changed to 0.001.  

\begin{figure}[t]
	\centering
	\includegraphics[width=0.99\linewidth]{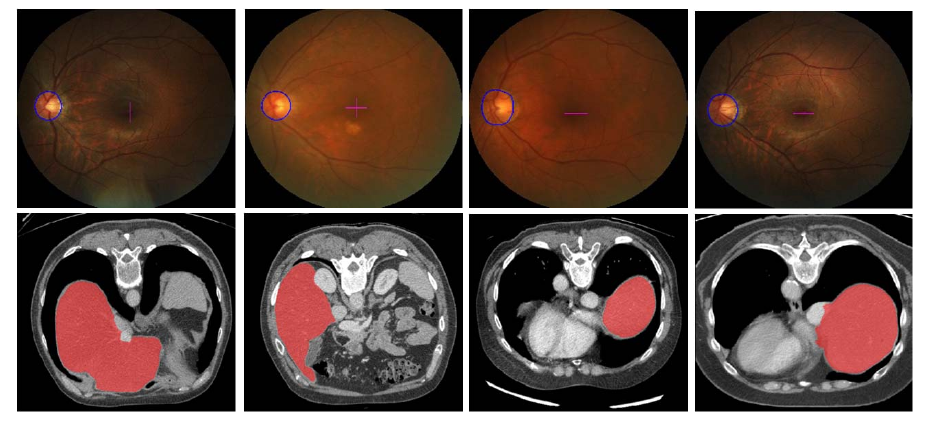}
	\vspace*{-4mm}
	\caption{Examples of our semi-supervised (20\%) segmentation results for the fundus image and liver CT scans. Blue color denotes the segmented boundary of optic disc and red color represents the segmented liver.}
	\label{fig:two}\centering
	\vspace*{-3mm}
\end{figure}

\revise{
We report \reviseagain{JA of the supervised} and semi-supervised results under the setting of 10\% labeled training images and 20\% labeled training images, respectively.\reviselast{
As shown in Table~\ref{tab:retina}, we also compare with the other semi-supervised methods.
It can be observed that our method achieves 1.82\% improvement under the 10\% labeled training setting, which ranked top among all these methods. \reviseagain{Also,} the improvement achieved by our method under the 20\% training setting is also the highest.} 
Figure~\ref{fig:two} shows some visual segmentation results of our semi-supervised method.}
We can see that our method can better capture the boundary of the optic disc structure.

\subsection{Experiments on LiTS dataset}
For this dataset, we evaluate the performance of liver segmentation from CT volumes. 
Under our semi-supervised setting, we randomly separated the original 131 training data from the challenge into 121 training volumes and 10 testing volumes. 
For image preprocessing, we truncated the image intensity values of all scans to the range of [-200, 250] HU to remove the irrelevant details.
We run experiments with 3D U-Net~\cite{cciccek20163d} to verify the effectiveness of our method.  For the 3D U-Net, the input size is randomly cropped to $112 \times 112 \times 32$ to leverage the information from the third dimension. 

According to the evaluation of \reviseagain{the 2017 LiTS challenge}, we employed Dice per case score to evaluate the liver segmentation result, which refers to an average Dice score per volume. 
We report the performance of our method and \reviseagain{the other three} semi-supervised methods \reviseagain{under the settings} of 10\% labeled training images and 20\% labeled training images, respectively, in Table~\ref{tab:lits}.\revise{
We can see that our approach achieves the highest performance improvement in both the 10\% and 20\% labeled training settings, with 5.33\% and 5.72\% improvements, respectively.\reviselast{
In semi-supervised learning, it is obvious that our method gains higher performance consistently than other methods ~\cite{zhang2017deep,bai2017semi,Hung_semiseg_2018,tarvainen2017mean} in both 10\% and 20\% settings, respectively.}
We also visualize some liver segmentation results from CT scans in the second row \reviseagain{of Figure~\ref{fig:two}}.
}

\section{Discussion}

Supervised deep learning has been proven extremely effective for many problems in \reviseagain{the medical image community}. 
However, promising performance heavily relies \reviseagain{on the amount of annotations.}
Developing new methods with limited annotation will largely advance\reviseagain{ real-world clinical applications}.  
In this work, we focus on developing semi-supervised learning methods for medical image segmentation,~\reviseagain{which have great} potential to reduce the annotation effort by taking advantage of \reviseagain{numerous unlabeled data}.
The key insight of our semi-supervised learning method is the transformation-consistent self-ensembling strategy.
Extensive experiments on three representative and challenging datasets \reviseagain{demonstrated the effectiveness} of our method.

Medical image data has different formats,~\eg, 2D in-plane scans (\eg, dermoscopy images and fundus images) and 3D volumetric data (\eg, MRI, CT).  
In this paper, we use both 2D and 3D networks for segmentation.
\reviselast{
Our method is flexible and can be easily applied to both 2D and 3D networks.
It is worth noting that recent works~\cite{zhou2018semi,xia20183d} are specifically designed for 3D volume data by considering three-view co-training, \ie, the coronal, sagittal and axial views of the \reviseagain{volume data}. 
However, we aim for a more general approach that is applicable \reviseagain{to 2D and 3D} medical images.  
For 3D semi-supervised learning, it could be a promising direction to design specific methods by considering the 3D natural property of the volumetric data.}

\reviseagain{Recent works on} network equivariance~\cite{worrall2017harmonic,cohen2016group,pmlr-v48-dieleman16} improve the generalization capacity of \reviseagain{trained networks} by exploring the equivariance property. \reviselast{
E.g.,~\citet{cohen2016group} presented a group equivariant neural network.
Our method also leverages the transformation consistency principle, but differently, we \reviseagain{aim for semi-supervised segmentation}.
Moreover, if we trained these works, \ie, harmonic network~\cite{worrall2017harmonic}, in a semi-supervised way to leverage unlabeled data, the transformation regularization will have no effect ideally, since the network outputs are the same when applying the transformation to the input images. 
Hence, the limited regularization would restrict the performance improvement from unlabeled data.}

One limitation of our method is that we assume both labeled and unlabeled data come from the same distribution. 
However, in clinical applications, \reviseagain{labeled and unlabeled data may have different distributions with a domain shift.} 
\citeauthor{oliver2018realistic}~\cite{oliver2018realistic} demonstrated that the performance of semi-supervised learning methods can degrade substantially when the unlabeled dataset contains out-of-distribution samples. 
However, most of the current semi-supervised approaches for medical image segmentation do not consider this issue.
Therefore, in the future, we would explore domain adaptation~\cite{kamnitsas2017unsupervised}, and investigate how to adopt it with \reviseagain{a self-ensembling strategy}.

\reviseagain{
In our work, the selection of transformation is based on the property of neural networks. The convolution layer is not rotation equivariant. To tackle the segmentation task, we need to train the network to be rotation equivariant. Moreover, the neural network is not scale equivariant due to padding, upsampling, etc. The rotation and scaling transformations are the general transformations used in medical images. Thus, learning to minimize the output differences caused by these transformations will regularize the network to be transformation-consistent.
Our method is flexible to extend to more general transformation cases, such as affine transformations. The transformation-consistent module consists of a transformation on the input image that will be fed to the teacher model, and \reviseagain{the same} transformation on the output space generated by the student model. 
It is flexible without additional training costs to be applied to the neural network. 
Moreover, recent automatic augmentation search works~\cite{cubuk2019autoaugment,lim2019fast,buslaev2020albumentations} explored the best transformations for a specific dataset.
It is an interesting future work to explore more useful transformations for our semi-supervised segmentation framework through automatic data augmentation. 
Also, the experiments reported are the averaged result over three trials, which also indicates the robustness of our method. 
 }

\section{Conclusion}
\reviseagain{This paper presents a novel and effective transformation-consistent self-ensembling semi-supervised method for medical image segmentation.
The whole framework is trained with a teacher-student scheme and optimized by a weighted combination of supervised and unsupervised losses.}
To achieve this, we introduce a transformation-consistent self-ensembling model for the segmentation task, enhancing the regularization and can be easily applied on 2D and 3D networks.~\reviseagain{Comprehensive experimental analysis on three medical imaging datasets, \ie, skin lesion, retinal image, and liver CT datasets, demonstrated the effectiveness of our method.}
Our method is general and can be widely used in other semi-supervised medical imaging problems.

\bibliographystyle{IEEEtranN}
\small{\bibliography{refs}}

\if 1
\begin{IEEEbiography}[{\includegraphics[width=1in,height=1.25in,clip,keepaspectratio]{photo/meng3.jpg}}]
	{Xiaomeng Li} is currently a postdoctoral research fellow in the Department of Radiation Oncology at Stanford University. She obtained Ph.D degree in the Department of Computer Science and Engineering, The Chinese University of Hong Kong (CUHK) in July 2019.
	Her research include medical image analysis, computer vision and deep learning.
	\vspace{-13mm}
\end{IEEEbiography}

\begin{IEEEbiography}[{\includegraphics[width=1in,height=1.25in,clip,keepaspectratio]{photo/quan.jpg}}]
	{Lequan Yu} received the B. Eng degree from Department of Computer Science and Technology in Zhejiang University in 2015.
	He obtained Ph.D degree in the Department of Computer Science and Engineering, The Chinese University of Hong Kong (CUHK) in July 2019 .
	He is currently a postdoctoral research fellow in the Department of Radiation Oncology at Stanford University.
	His research lies at the intersection of medical image analysis and artificial intelligence.
		\vspace{-13mm}
\end{IEEEbiography}

\begin{IEEEbiography}[{\includegraphics[width=1in,height=1.25in,clip,keepaspectratio]{photo/hao.jpg}}]
{Hao Chen} received the B. Eng degree in School of Information Engineering from Beihang University (BUAA).
He obtained Ph.D degree in the Department of Computer Science and Engineering, The Chinese University of Hong Kong (CUHK) in July 2017. He focuses on translating advanced deep learning techniques into real-world clinical products.
	\vspace{-13mm}
\end{IEEEbiography}

\begin{IEEEbiography}[{\includegraphics[width=1in,height=1.25in,clip,keepaspectratio]{photo/philip.png}}]
{Chi-Wing Fu} is currently an associate professor in the Chinese University of Hong Kong.  He served as the co-chair of SIGGRAPH ASIA 2016's Technical Brief and Poster program, associate editor of Computer Graphics Forum, and panel member in SIGGRAPH 2019 Doctoral Consortium, as well as program committee members in various research conferences, including SIGGRAPH Asia Technical Brief, SIGGRAPH Asia Emerging tech., IEEE visualization, CVPR, IEEE VR, VRST, Pacific Graphics, GMP, etc.  His recent research interests include computation fabrication, 3D computer vision, user interaction, and data visualization. 
\vspace{-13mm}
\end{IEEEbiography}

\begin{IEEEbiography}[{\includegraphics[width=1in,height=1.25in,clip,keepaspectratio]{photo/lei.jpg}}]
{Lei Xing} is currently the Jacob Haimson Professor of Medical Physics and Director of Medical Physics Division of Radiation Oncology Department at Stanford University.  He obtained his PhD in Physics from the Johns Hopkins University in 1992. He has been a member of the Radiation Oncology faculty at Stanford since 1997. His research has been focused on inverse treatment planning, artificial intelligence (AI) in medicine, tomographic image reconstruction, CT, optical and PET imaging instrumentations, image guided interventions, nanomedicine, and applications of molecular imaging in radiation oncology. 
\vspace{-13mm}
\end{IEEEbiography}

\begin{IEEEbiography}[{\includegraphics[width=1in,height=1.25in,clip,keepaspectratio]{photo/heng.png}}]
{Pheng-Ann Heng} received his B.Sc. (Computer Science) from the National University of Singapore in 1985. 
He received his M.Sc. (Computer Science), M. Art (Applied Math) and Ph.D. (Computer Science) all from the Indiana University in 1987, 1988, 1992 respectively.
He is a professor at the Department of Computer Science and Engineering at The Chinese University of Hong Kong. He has served as the Department Chairman from 2014 to 2017 and as the Head of Graduate Division from 2005 to 2008 and then again from 2011 to 2016. 
His research interests include AI and VR for medical applications, surgical simulation, visualization, graphics and human-computer interaction.
\vspace{-13mm}
\end{IEEEbiography}

\fi 
\ifCLASSOPTIONcaptionsoff
\newpage

\fi\end{document}